\documentclass{article}

\PassOptionsToPackage{numbers, compress}{natbib}
 \usepackage{graphicx}
 \usepackage[dblblindworkshop, final]{neurips_2025}
\usepackage[utf8]{inputenc} % allow utf-8 input
\usepackage[T1]{fontenc}    % use 8-bit T1 fonts
\usepackage{hyperref}       % hyperlinks
\usepackage{url}            % simple URL typesetting
\usepackage{booktabs}       % professional-quality tables
\usepackage{amsfonts}       % blackboard math symbols
\usepackage{nicefrac}       % compact symbols for 1/2, etc.
\usepackage{microtype}      % microtypography
\usepackage{xcolor}         % colors
\usepackage{listings}
\usepackage{float}
\usepackage{natbib}
\usepackage{markdown}

\title{A Matter of Representation: Towards Graph-Based Abstract Code Generation}
\workshoptitle{Deep Learning for Code}

\author{%
  Nyx Iskandar \\
  University of California, Berkeley \\
  \texttt{nyx@berkeley.edu} \\
  \AND
  Hisham Bedri \\
  Ramen VR \\
  \texttt{hisham@ramenvr.com} \\
  \And
  Andy Tsen \\
  Ramen VR \\
  \texttt{andy@ramenvr.com} \\
}

\begin{document}

\maketitle

\begin{abstract}
  Most large language models (LLMs) today excel at generating raw, sequential code with minimal abstractions and custom structures. However, there has been little work on graph-based abstract code generation, where significant logic is encapsulated in predefined nodes and execution flow is determined by edges. This is relevant for visual programming languages, and in cases where raw source code is inaccessible to users and LLM training sets. In this work, we propose and evaluate JSON representations for graphs to enable high accuracy graph-based abstract code generation. We evaluate these representations on ScratchTest, a mini-benchmark based on our custom Python re-implementation of Scratch, which tests the LLM in code graph space. Our findings demonstrate that LLMs can indeed perform the aforementioned generation task in a single pass without relying on specialized or complex pipelines, given the correct graph representations. We also show that different representations induce significantly different accuracies, highlighting the instrumental role of representations in this generation task. All in all, this work establishes the first steps towards representation learning for graph-based abstract code generation.
\end{abstract}

\section{Introduction}

Large language models (LLMs) \citep{brown2020languagemodelsfewshotlearners} have increased in capability significantly over recent years in not only natural language tasks, but also the more syntactically rigorous task of code generation \citep{chen2021evaluatinglargelanguagemodels, athiwaratkun2023multilingualevaluationcodegeneration}. These models excel not only in code completion \citep{guo2023longcoderlongrangepretrainedlanguage, wang2021codecompletionmodelingflattened, wu2024repoformerselectiveretrievalrepositorylevel}, but also in instruction-following to generate code in natural-language-to-code (NL2Code) tasks \citep{athiwaratkun2023multilingualevaluationcodegeneration, austin2021programsynthesislargelanguage, zan2023largelanguagemodelsmeet}. Indeed, many LLMs today, be they open-weight or otherwise, feature a code fine-tuning of them, such as Qwen3-Coder \citep{yang2025qwen3technicalreport, qwen3_coder}, or have their coding capabilities highlighted by model developers and evaluators through blog highlights and benchmark performance reporting, such as those for GPT-5 \citep{gpt5}, Claude Opus 4.1 \citep{claude_opus_41_2025}, and DeepSeek-R1 \citep{deepseekr1}. This is in no small part due to the abundance of benchmarks that specifically evaluate the code generation, understanding, and/or problem-solving capabilities of these models, such as SWE-bench Verified \citep{swe_bench_verified} and Weapons of Mass Destruction Proxy - Cyber \citep{li2024wmdpbenchmarkmeasuringreducing}. The success of agentic tools in Cursor \citep{cursor}, Codex \citep{codex}, and ClaudeCode \citep{claudecode} is a testament to how capable these models are in coding, and how rapidly these capabilities have developed in the past two years of their massive popularity and widespread adoption.

However, these coding tasks are restricted to the domain of raw, sequential code, which is the style of code that software engineers are most familiar with and thus have produced an abundance of data of that is available online for training: neatly arranged line-by-line in a file as a string in an intuitively linear fashion. Of course, code need not necessarily be written in this form; there are many graph-based languages like Scratch \citep{scratch}, Unreal Engine Blueprints \citep{blueprints_ue}, and n8n software \citep{n8n_2025} that are widely used by beginner programmers for learning how to think programmatically, game developers and designers for rapid prototyping, and software engineers for implementing agentic AI workflows, respectively. Despite its prevalence, the domain of graph-based abstract code has thus far been sidelined, which is evident in the lack of benchmarks related to it. There is also no unified representation of these graph-based code fragments. As such, LLMs today lack the ability to reliably generate logically sound graph-based code that accomplishes the task or goal detailed in the user query, or even graph-based code that compiles.

Furthermore, research on graph-based abstract code generation has greater implications in supporting LLMs in cases where there are minimal code implementation examples available, such as in newer libraries. This is because inherent in this flavor of code is the fact that implementation details are abstracted away from the user and thus the LLM. While it can be argued that any programming language abstracts implementation details away, the abundance of available data of such abstractions in, for instance, GitHub repositories that is exposed to these LLMs during pre-training and fine-tuning eliminates that barrier \citep{kocetkov2022stack3tbpermissively, lozhkov2024starcoder2stackv2}. However, for lesser-known or newer libraries or frameworks, this is hardly the case. In fact, we argue that these libraries or frameworks face the same challenge as graph-based coding languages: abstraction. Furthermore, the existence of Scratch, Blueprints, and other similar languages shows that raw code can be represented as graphs, and thus the code underlying these lesser-known or newer libraries or frameworks can also be represented as graphs. While in-context learning \citep{icl} seems like a potential solution, it is more suited for familiarizing LLMs with new syntax, not logic flows; graph-based languages introduce interesting non-linear relationships between code fragments in the form of nodes, posing a challenge in-context learning is ill-equipped to solve. As such, advancing graph-based code generation also advances the general capability of LLMs to generate code using the libraries or frameworks that are disproportionately underrepresented in their training data.

As such, \textbf{our contributions} are as follows:
\begin{itemize}
    \item We propose a simple JSON representation of \textbf{nodes} that enables current LLMs to generate the most syntactically and logically accurate code graphs when given a list of those representations as \textbf{reference} compared to other JSON representations.
    \item We propose a JSON representation of \textbf{code graphs} that further enables current LLMs to generate the most syntactically and logically accurate code graphs when \textbf{outputting} that representation compared to other JSON representations.
    \item We propose a new \textbf{mini-benchmark} based on a Python re-implementation of Scratch to evaluate graph-based abstract code generation capabilities of LLMs.
\end{itemize}

Our work focuses on improving the performance of a one-agent LLM framework without fine-tuning \citep{luo2025wizardcoderempoweringcodelarge, dettmers2023qloraefficientfinetuningquantized}, in-context learning \citep{icl}, or other additions to the base model. In other words, we focus on showing how integrating the correct node and graph representations into the LLM's prompt increases the LLM's generation accuracy. Code is available in the \href{https://github.com/xyntechx/graph-abstract-code-gen}{GitHub repository}.

\section{Related Work}

\subsection{Graph Understanding and Generation}
Past work on graph generation includes GraphRNN \citep{you2018graphrnngeneratingrealisticgraphs} and GraphGAN \citep{wang2017graphgangraphrepresentationlearning}. Though leveraging distinct approaches, both are suited for generating generic nodes and edges by learning a particular graph distribution represented in the ground truth set. However, these approaches do not translate well for our focus on graph-based code generation; these past approaches focus on instantiating graph models based on characteristics or properties of a set of observed graphs, while our focus is on generating graphs which fulfill a functional objective as we work with rich nodes that encode information and rich edges that encode logic and data flow. The same limitation exists in \citet{fatemi2023talklikegraphencoding}, \citet{wang2024languagemodelssolvegraph}, \citet{ye2024languagegraphneeds}, and \citet{zhang2023graphtoolformerempowerllmsgraph} as the foci of these papers are on graphs as mathematical objects instead of representations of logic, and on evaluating LLM graph understanding (e.g. whether two nodes are connected to each other) instead of on completion based on the encoded logic within each graph.

More recently, there has been work on Graph Foundation Models (GFMs) \citep{graphfoundationmodels}, which are built on either graph neural networks (GNNs) \citep{gnn} or LLMs. While GFMs are designed to handle non-Euclidean data for graphs, we argue that it is unnecessary, at least for our focus, as graphs can be represented as plain string or stringified JSON, both being Euclidean. As to the backbones of GFMs, it has been shown that GNNs do not scale well with current hardware \citep{empiricalanalysisofperformancebottlenecksingnn, onthescalabilityofgnnsformoleculargraphs, largescalegnnaaai}, while the representation used by the proposed LLM-based GFMs relies too heavily on brittle natural language rather than structured representations like JSON. The limitations of GFMs also apply to the work by \citet{jin2024largelanguagemodelsgraphs} as they leverage GNNs and/or natural language translations of graphs. Instead, we show that with a correct structured representation of code graphs, vanilla LLMs alone suffice in generating syntactically and logically sound graphs.

\subsection{LLM Coding Agents}
The other side of our work is code generation, which today's LLMs are extremely capable of, as aforementioned. Recent work in reinforcement learning on chain-of-thought has increased LLM capabilities in verifiable domains, including mathematics and coding \citep{wei2023chainofthoughtpromptingelicitsreasoning, yeo2025demystifyinglongchainofthoughtreasoning, deepseekr1}. Prior to that, methods like fill-in-the-middle (FIM) \citep{bavarian2022efficienttraininglanguagemodels} have proved effective in training code models like CodeGemma \citep{codegemmateam2024codegemmaopencodemodels}, and data processing methods in the form of file-level and repo-level pre-training have also proven effective in increasing the accuracy of code models like Qwen2.5-Coder \citep{hui2024qwen25codertechnicalreport}. LLM code capabilities have also been enhanced by constrained decoding for structured output generation \citep{guidanceai, dong2025xgrammarflexibleefficientstructured, willard2023efficientguidedgenerationlarge, ggml_llamacpp}, as much of software engineering relies on reliable structures for reliable data flows.

However, it is evident that LLMs, be they foundation models or fine-tunes or agents, show greater capability in generating certain languages, libraries, frameworks, etc. than others \citep{peng2024humanevalxlmultilingualcodegeneration, athiwaratkun2023multilingualevaluationcodegeneration, patil2023gorillalargelanguagemodel}. This discrepancy is due to the different levels of availability of data online for different domains. We thus hypothesize that having LLMs generate graph-based code in an ubiquitous format like JSON will increase their accuracy.

\section{ScratchTest}
Before discussing our proposed representations, it is valuable to first explain the mini-benchmark we used to run the evaluations, and to ground what we mean by graph-based abstract code in real examples. This section thus discusses ScratchTest, a set of test prompts and correct behaviors that assesses the accuracy of an LLM in implementing nodes of a re-implementation of Scratch \citep{scratch} written in Python.

\begin{figure}[H]
  \centering
  \includegraphics[scale=0.5]{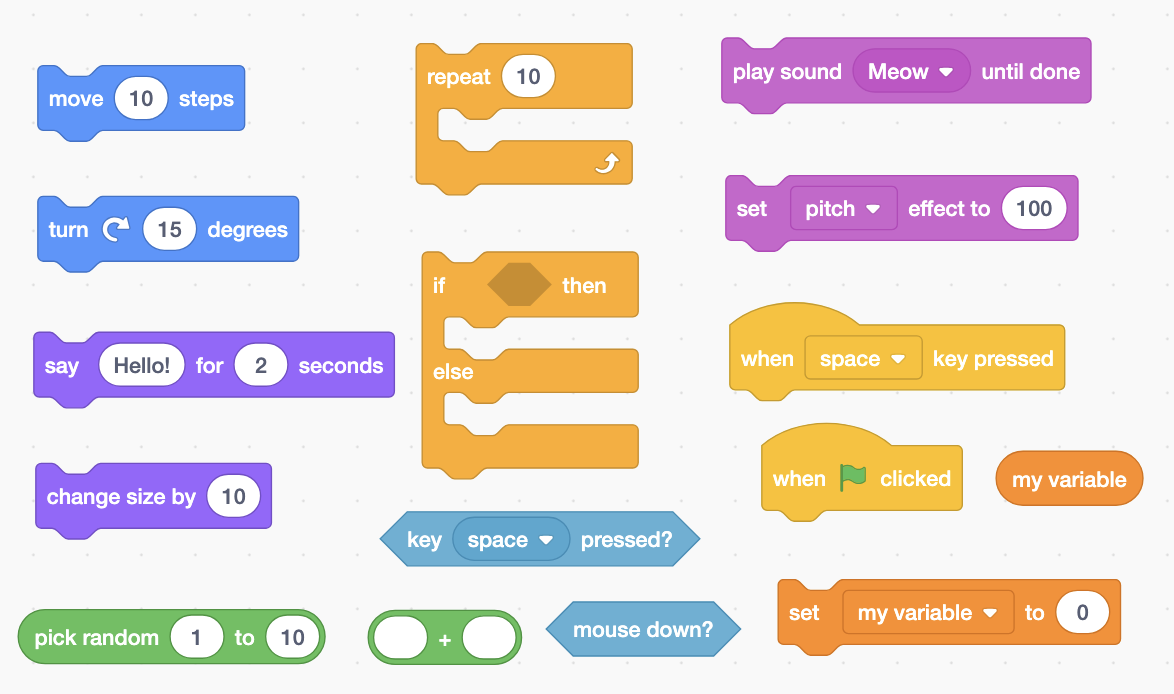}
  \caption{Two Scratch blocks from each built-in block type in the Scratch online editor.}
  \label{fig:scratch-blocks}
\end{figure}

Scratch is a beginner-friendly, high-level programming language that is based on the concept of blocks, which can be thought of as nodes in a graph (we thus use ``block'' and ``node'' interchangeably in this paper). Programs in Scratch primarily manipulate a sprite using built-in blocks of various types: motion, looks, sound, event, control, sensing, operator, and variable \citep{scratch}. A subset of these blocks is found in Figure \ref{fig:scratch-blocks}. Scratch blocks abstract away many implementation details, hence are used by users through connecting them together (as indicated by the indents and outdents of each block) and modifying the default parameters (e.g. 10 in ``move 10 steps''), rather than writing lines of code that are more akin to Python or C++ to manipulate primitives and keywords. ScratchTest includes Python re-implementations of 53 out of the 107 built-in Scratch blocks, chosen by which blocks can be feasibly implemented in a command line interface (CLI). The full list of blocks that are included in ScratchTest is found in Appendix \ref{a:all-blocks}.

To preserve simplicity in implementation and output analysis, and given the constraints of a CLI, the blocks in ScratchTest do not directly manipulate a sprite nor any visual element. Instead, we rely on each block returning a behavior log (e.g. a ``move n steps'' block would output ``moved n steps'' when called). We also persist a dictionary of sprite attributes, such as its x-position, y-position, and direction (i.e. rotation). Furthermore, each complete and valid graph as the final output of our proposed methodology outputs a Python file that directly interfaces with the classes and functions of the Scratch re-implementation. As such, assessing both the behavior logs outputted by the graphs generated by LLMs and the output Python file is how we evaluate the accuracy of these LLMs.

Note that our mini-benchmark currently does not have a canonical set of ground-truth behavior logs nor Python files due to the high variability in approaches to satisfy the functionality requested in a prompt. While we can curate more straightforward prompts to circumvent this issue, this risks asking the LLM to implement functionalities which are too elementary, preventing the tests from being sensitive enough to evaluate the proposed representations and their ablations or alternatives. Hence, to ensure reliability and accuracy in the evaluation, we opt to manually evaluate both the behavior logs and Python files of sufficiently complex prompts. A correct implementation is one which outputs a reasonable behavior log and a logically accurate Python file based on the requested functionality in the prompt.

\begin{figure}[H]
  \centering
  \includegraphics[scale=0.28]{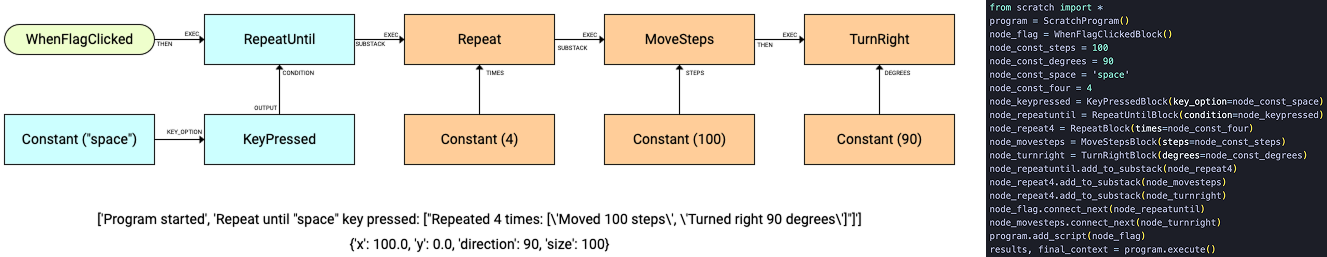}
  \caption{Correct output graph, behavior log, sprite state, and corresponding Pythonic Scratch for prompt ``When the green flag is clicked, continuously move in a square pattern until the user presses the space key.''}
  \label{fig:sample-graph}
\end{figure}

ScratchTest includes 20 unique prompts that describe a certain functionality; for example, ``When the green flag is clicked, continuously move in a square pattern until the user presses the space key.'' and ``When the `s' key is pressed, say a secret password made of two random letters and three random numbers.'' Each prompt's requested functionality can be accomplished by implementing at least one permutation of the nodes available in ScratchTest. Figure \ref{fig:sample-graph} shows a sample correct graph of Scratch nodes (with connected ports specified) with the corresponding behavior log, sprite state, and Python file for the prompt specified in the caption. Each prompt is run five separate times; the resulting behavior log and Python file corresponding to each prompt are thus also evaluated five separate times.

\section{Representations}
This section explains our proposed JSON representations to ensure the highest accuracy in graph generation for a vanilla one-agent LLM framework compared to other JSON representations. These include the node representation we provide the LLM as reference, as well as the graph representation we steer the LLM to output. We include a \textbf{reference node representation} to provide the LLM with context on what nodes (i.e. functions) it can use, which is crucial as it is not fine-tuned on the particular graph-based library or framework. This is distinct from in-context learning; performing in-context learning would be providing the LLM with correct prompt-graph pairs. We also include an \textbf{output graph representation} to guide the LLM to generate a standardized structured output format for the post-processing step that translates the generated graph into its equivalent in the Pythonic Scratch implementation.

For our main experiments, we use OpenAI's gpt-oss-120b \citep{gpt_oss} on Groq \citep{groq_2025} as is. The exact model configuration is found in Appendix \ref{a:llm-config}. We also ran experiments on other LLMs, namely qwen3-32b \citep{yang2025qwen3technicalreport}, deepseek-r1-distill-llama-70b \citep{deepseekr1}, and llama-3.3-70b-versatile \citep{grattafiori2024llama3herdmodels} on Groq; their respective configurations are also found in Appendix \ref{a:llm-config}. The system prompt can be found in Appendix \ref{a:llm-system}.

It is important to note that the LLM is exposed only to the representations detailed in this section. Notably, the Python implementation of Scratch is not at all provided to the LLM, which is to say that the LLM is not aware of what logic underlies the JSON representations it is actually provided with. This simulates real-world environments where the user (or LLM) is only given the function names without the underlying logic that has been abstracted away. As such, the LLM is unable to use or infer from the actual implementation to assist its generation, which is desired.

\subsection{Reference Node Representation} \label{repr:node}
We propose a node representation with the following format to be provided to the LLM as a reference list of canonical nodes:

\lstset{
  basicstyle=\footnotesize,
  xleftmargin=.2\textwidth, xrightmargin=.2\textwidth
}
\begin{lstlisting}
[NODENAME]: {
    inPorts: {id: string, type: string}[],
    fields: {id: string, type: string}[],
    outPorts: {id: string, type: string}[]
}
\end{lstlisting}

More specifically:
\begin{itemize}
    \item \textbf{NODENAME}: The name of the node. Each name corresponds to that of the actual block in Scratch (e.g. WhenFlagClicked corresponds to the ``when green flag clicked'' block in Scratch).
    \item \textbf{inPorts}: The input ports of the node, which are ports that correspond to an incoming edge. These include ports that represent parameters, and EXEC ports which are connected to an edge that is also connected to the node that immediately precedes the current node in execution (the visual indent of a Scratch block is EXEC).
    \item \textbf{fields}: Parameters of the node that must be chosen from a predefined list. For instance, the MathFunction node has a field called OPERATOR which takes in only the following values: abs, floor, ceiling, sqrt, sin, cos, tan, asin, acos, atan, ln, log, e\textasciicircum, 10\textasciicircum. These are virtually indistinguishable from \textbf{inPorts} during generation.
    \item \textbf{outPorts}: The output ports of the node, which are ports that correspond to an outgoing edge. These include ports that represent return values, THEN ports which are connected to an edge that is also connected to the node that immediately succeeds the current node in execution (the visual outdent of a Scratch block is THEN), and SUBSTACK ports which are connected to edges that are also connected to the nodes that exist within the loop/control block (only relevant for control nodes).
    \item \textbf{id}: The unique identifier of each port (e.g. EXEC).
    \item \textbf{type}: The type that is to be connected to this port (e.g. number, string, Node).
\end{itemize}

We find that the addition of \textbf{type} to each port increases the accuracy of the graph generation, as we observe that the LLM may hallucinate node-node or constant-node connections of incompatible types (e.g. connecting a port of type number to a port of type string). Interestingly, we find that adding natural language descriptions to both the nodes and the ports does not increase the accuracy of the LLM in constructing the graph by any significant margin. We hypothesize that this is because the node and port names are sufficiently descriptive for the LLM such that the addition of these descriptions only increases the token count for a comparable accuracy. We will show in Section \ref{evaluation} the quantitative effects of these additions on the accuracy of the generation through ablations.

The inclusion of \textbf{id} for each port helps in the post-processing step that translates the output graph generated by the LLM to the equivalent Pythonic Scratch implementation. This is because each port can then be uniquely identified and thus be well-assembled after the translation. Port IDs generally correspond to parameter names (e.g. STEPS for the number of steps to move in a MoveSteps nodes), but we also have the specially defined ports THEN and EXEC to facilitate the reconstruction of execution flow such that the THEN port of some node is connected to the EXEC port of the next node in the main graph. Another specially defined port is SUBSTACK, which is relevant only for control nodes (e.g. Repeat, IfElse); the SUBSTACK port of a control node is connected to the EXEC port of the nodes to be repeated, looped over, branched into, etc. Some nodes like IfElse have multiple SUBSTACK ports to handle logic branches (i.e. one for the case where the condition is true, and another for the case where the condition is false).

\subsection{Output Graph Representation} \label{repr:graph}
We propose a graph representation with the following format to be outputted by the LLM:

\lstset{
  basicstyle=\footnotesize,
  xleftmargin=.2\textwidth, xrightmargin=.2\textwidth
}
\begin{lstlisting}
{
    nodes: {
        [key: string]: {
            name: string,
            value: any|null
        }
    },
    edges: {
        outNodeID: string,
        outPortID: string,
        inNodeID: string,
        inPortID: string
    }[]
}
\end{lstlisting}

More specifically:
\begin{itemize}
    \item \textbf{nodes}: A map of \textbf{nodeIDs} to node data. nodeIDs are unique string identifiers such that no two nodes in one graph can have the same ID. The node data dictionary has the node name and the node value. The value field is always $null$ except for when it is for a constant, in which case its value is the value it is supposed to hold (e.g. 10, ``Hello'', True). Note that unique nodeIDs are necessary as nodes of the same name may be instantiated more than once in one graph (e.g. the graph may have multiple Add nodes).
    \item \textbf{edges}: An array of dictionaries representing edges. \textbf{outNodeID} belongs to the node whose outPort is used for this edge (\textbf{outPortID} is the ID of said port). \textbf{inNodeID} belongs to the node whose inPort is used for this edge (\textbf{inPortID} is the ID of said port). A \textbf{field} of a node can be considered an inPort for the purposes of an edge.
\end{itemize}

As will be shown in our experiments detailed in the subsequent section, this particular output graph representation achieves the highest generation accuracy compared to an alternative representation which is equally as intuitive. We hypothesize that the separation of concerns between generating the nodes and the edges reduces the margin of error for the LLM, and the generation itself is more linear in the sense that connectable nodes are defined first and edges connecting those nodes are defined later. This is in contrast to a representation where nodes and edges are defined within the same atomic JSON object, where nodes that are defined later are referenced earlier, and one edge must be defined twice (first from the perspective of the outNode via the outPin, and second from the perspective of the inNode via the inPin), resulting in duplicative and more error-prone effort.

The proposed output graph representation will then undergo a post-processing step to translate the graph into the Python implementation of Scratch. The reconstruction first orders the nodes and edges using topological sort, as the graph must be a directed acyclic graph for it to have a valid execution sequence; the existence of for loops and other similar logic is handled by the respective nodes, which prevents instances of actual loops in the graph representation. After the sort, the post-processing step constructs a Python file which contains the Pythonic Scratch methods obtained after translating the nodes and edges in the graph representation. The translation process includes instantiating Scratch block objects bound to variables named the respective nodeIDs, defining constants, passing in nodes or constants as arguments to the block objects whenever necessary, adding nodes to the respective substacks of control nodes, and connecting the nodes sequentially based on the THEN and EXEC ports.

\section{Evaluation} \label{evaluation}

We conduct ablation studies to evaluate the impact of including and omitting certain components from both the reference node representation and the output graph representation. As a reminder, a correct implementation is one that outputs a reasonable behavior log and a logically accurate Python file; an incorrect implementation is either logically flawed or results in an error during the post-processing step or during the execution of the constructed Pythonic Scratch implementation. The results shown in Sections \ref{evaluation:node} and \ref{evaluation:graph} are for gpt-oss-120b; results for other models can be found in Appendix \ref{a:other-llms}.

\subsection{Reference Node Representation} \label{evaluation:node}

Breaking down the ablations, we have the following three reference node representations:
\begin{itemize}
    \item \textbf{No Types}: Our proposed representation, but without types for both nodes and ports. This representation is essentially the minimal baseline representation.
    \item \textbf{Extra Description}: Our proposed representation, but with description fields for both nodes and ports (e.g. the description for the MoveSteps node is ``Move sprite forward by specified number of steps'', and that for the EXEC port is ``Previous block that triggers this block'').
    \item \textbf{Proposed}: Our actual proposed representation described in Section \ref{repr:node}.
\end{itemize}

Evaluating the ablations on 20 ScratchTest prompts five independent times each, we obtain the results shown in Table \ref{table:node-ablations}. For a granular per-prompt breakdown of the results, please refer to Appendix \ref{a:rnbreakdown}.

\begin{table}[H]
  \caption{ScratchTest results for reference node representation ablations.}
  \label{table:node-ablations}
  \centering
  \begin{tabular}{lrrrrr}
    \toprule
    Name                  & Run 1  & Run 2 & Run 3 & Run 4 & Run 5 \\
    \midrule
    No Types              & 12/20            & 14/20           & 12/20           & 15/20           & 12/20   \\
    Extra Description     & 14/20            & \textbf{16}/20  & 14/20           & \textbf{17}/20  & 13/20   \\
    \textbf{Proposed}     & \textbf{16}/20   & 14/20           & \textbf{15}/20  & 16/20           & \textbf{14}/20   \\
    \bottomrule
  \end{tabular}
\end{table}

We conclude that \textbf{Proposed} has the highest mean accuracy (albeit only slightly more accurate than \textbf{Extra Description}) and is the most consistent across runs. The results summary is shown in Table \ref{table:node-ablations-summary}.

\begin{table}[H]
  \caption{Summary results for reference node representation ablations.}
  \label{table:node-ablations-summary}
  \centering
  \begin{tabular}{lcc}
    \toprule
    Name                  & Mean  & Standard Deviation \\
    \midrule
    No Types              & 0.65          & 0.071  \\
    Extra Description     & 0.74          & 0.082   \\
    \textbf{Proposed}     & \textbf{0.75} & \textbf{0.050}    \\
    \bottomrule
  \end{tabular}
\end{table}

Calculating the two-sided p-values with Welch's t-test, and setting the significance level $\alpha = 0.05$, the difference in accuracy between \textbf{Proposed} and \textbf{No Types} is significant as $p=0.036 \leq \alpha$; however, that between \textbf{Proposed} and \textbf{Extra Description} is not significant as $p=0.82 > \alpha$. This shows that while adding types increases accuracy by a statistically significant amount, adding descriptions does not; descriptions are unnecessary, cost more in terms of tokens, and do not scale as robustly due to the extra tokens per node. We posit that descriptions must be well-thought out and sufficiently verbose to result in a significant increase in accuracy (but at the cost of tokens) as otherwise accuracy may instead decrease due to misleading or vague descriptions; a more scalable alternative is to name the nodes and ports sufficiently descriptively, and if necessary create a mapping between the more descriptive names to the actual names provided by the abstracted library or package. The exact descriptions used are all included in the \href{https://github.com/xyntechx/graph-abstract-code-gen}{GitHub repository}, where one can see the difference in token counts between \textbf{Proposed} and \textbf{Extra Description}.

\subsection{Output Graph Representation} \label{evaluation:graph}

Breaking down the ablations, we have \textbf{Proposed}, which is our actual proposed representation described in Section \ref{repr:graph} (this is identical to \textbf{Proposed} in Table \ref{table:node-ablations}), and \textbf{Alternative}, which is further described below:

\lstset{
  basicstyle=\footnotesize,
  xleftmargin=.2\textwidth, xrightmargin=.2\textwidth
}
\begin{lstlisting}
{
    [key: string]: {
        nodeName: string,
        value: any|null,
        edges: {
            portID: string,
            otherNodeID: string
        }[]
    }
}
\end{lstlisting}

More specifically (refer to Section \ref{repr:graph} for more explanation of the respective attributes):
\begin{itemize}
    \item \textbf{key}: The keys of the JSON object are \textbf{nodeIDs}.
    \item \textbf{nodeName}: The name of the node.
    \item \textbf{value}: The value of the node.
    \item \textbf{edges}: An array of dictionaries representing edges. \textbf{portID} is the ID of the port on this current node that is connected to this edge. \textbf{otherNodeID} is the ID of the other node that is connected to this edge (i.e. is paired with this current node by this edge).
\end{itemize}

Given the different output graph representation, the system prompt for \textbf{Alternative} is also different. It is outlined in Appendix \ref{a:llm-system-alt}.

Evaluating the ablations on 20 ScratchTest prompts five independent times each, we obtain the results shown in Table \ref{table:graph-ablations}. For a granular per-prompt breakdown of the results, please refer to Appendix \ref{a:ogbreakdown}.

\begin{table}[H]
  \caption{ScratchTest results for output graph representation ablations.}
  \label{table:graph-ablations}
  \centering
  \begin{tabular}{lrrrrr}
    \toprule
    Name                  & Run 1  & Run 2 & Run 3 & Run 4 & Run 5 \\
    \midrule
    Alternative           & 10/20            & 9/20                     & 10/20           & 11/20  & 9/20   \\
    \textbf{Proposed}     & \textbf{16}/20   & \textbf{14}/20           & \textbf{15}/20  & \textbf{16}/20           & \textbf{14}/20   \\
    \bottomrule
  \end{tabular}
\end{table}

The mean accuracy of \textbf{Alternative} is $0.49$ and its standard deviation is $0.042$; \textbf{Proposed} has a significantly higher mean accuracy and a slightly lower consistency (refer to Table \ref{table:node-ablations-summary}). Calculating the two-sided p-value with Welch's t-test, and setting the significance level $\alpha = 0.05$, the difference in accuracy between \textbf{Proposed} and \textbf{Alternative} is significant as $p=0.000024 \leq \alpha$. With $95\%$ confidence, \textbf{Proposed} improves accuracy by $23$–$29\%$ compared to \textbf{Alternative}.

Common errors faced by the alternative representation include connecting ports of the same direction together (i.e. connecting an outPort with another outPort, or an inPort with another inPort) and forgetting to define the ports for the two nodes that are connected by the particular edge (recall how \textbf{Alternative} requires one edge to be defined twice). These are fundamental and thus critical errors. Hence, we conclude that our proposed output graph representation indeed enables the LLM to generate more accurate graphs.

\section{Limitations}

Our proposed representations do not enable the LLM to achieve 100\% accuracy. We primarily observe these two common errors: failing to create a valid directed acyclic graph, and referencing an undeclared variable. It is to be noted that this work establishes the first steps towards graph-based abstract code generation, and we encourage and will ourselves embark on further research to find more effective representations and/or tokenization and encoding methods to increase the accuracy of the generated code graphs. Furthermore, we intentionally limit this work to the vanilla one-agent framework, recognizing that fine-tuned models as well as multi-agent frameworks that include planning and error-correction will likely boost the accuracy, especially for more complex long-horizon tasks, prompts which are exceedingly high-level (e.g. create a platformer game using only Scratch blocks), or frameworks or libraries with hundreds or thousands of nodes. Our goal in this paper is to propose representations that outperform other similar representations to act as a baseline for further refinement to the model or pipeline architecture.

\section{Conclusion}

It has been observed that the code generation capability of LLMs is a research focus prioritized by leading labs and companies \citep{chen2021evaluatinglargelanguagemodels, athiwaratkun2023multilingualevaluationcodegeneration, guo2023longcoderlongrangepretrainedlanguage, claudecode, qwen3_coder}, most advancing these capabilities through post-training methods like chain-of-thought reasoning for foundation models \citep{deepseekr1}, or relying on multi-turn methods leveraging planning and/or error-correction \citep{erdogan2025planandactimprovingplanningagents, kim2024llmcompilerparallelfunction} for downstream applications. However, the current focus is heavily on raw, sequential code with minimal abstractions, while minimal work has been done on graph-based abstract code. We argue that the latter is also an important direction in code generation, as not only do many software tools today leverage visual graph-based programming languages \citep{scratch, blueprints_ue, n8n_2025}, but also this flavor of code essentially represents niche or specialized libraries or frameworks that abstract away implementation details from downstream users and thus from training data for LLMs \citep{peng2024humanevalxlmultilingualcodegeneration, athiwaratkun2023multilingualevaluationcodegeneration, patil2023gorillalargelanguagemodel}. Our work acts as an early step towards graph-based abstract code generation, and we posit that finding the optimal representations of these graphs allows even current LLMs that are not specially trained to understand and generate non-sequential graphs to achieve a relatively high accuracy in this task.

A key idea of our proposed representations, both for reference nodes and output graphs, is that current LLMs are already frequently exposed to and thus well-capable of generating JSON or structured outputs. The question we aim to answer is thus ``what JSON representation will yield the highest accuracy consistently for graph-based abstract code generation using today's LLMs?'' Through evaluating ablations of our proposed representations on our new mini-benchmark ScratchTest, we conclude that our proposed representations are indeed the JSON representations that would yield the highest accuracy in the most consistent and resource-conservative manner, particularly for the one-agent framework without fine-tuning, in-context learning, multi-turn interactions, etc. Adding the aforementioned techniques and using frontier models will likely further increase the generation accuracy; what is valuable from our research for this future work is the baseline accuracy guarantees as demonstrated in the ScratchTest results. Given that our work proposes structures for reference nodes and output graphs, it can easily be adapted to larger and more complex graph-based domains beyond Scratch, as long as the context window of the LLM can accommodate the representations.

Furthermore, in attempting to answer the aforementioned question, we observe a behavior that has implications beyond graph-based abstract code generation. We learn that the capabilities of an LLM depend not only on its architecture and training data, but also the representation of information that it is provided within its context window during inference, and the representation of the output it is steered to generate. In other words, the same LLM can generate outputs of varying qualities given the same prompt structure but different representations. While we do not have definitive conclusions nor hypotheses as to why this occurs for the general case beyond the particular ablations evaluated in this work, it is nevertheless an important idea to bear in mind when working with these and similar models.

All in all, we hope that our work will spur more research into graph-based abstract code generation, either through searching for even better representations (that may not necessarily be JSON-based) or through innovating new algorithms and generative models that specialize in code graph space. We believe that these are promising directions for future work in deep learning for code.

\textbf{Acknowledging societal impacts}: We acknowledge that there are broader societal impacts of our work. Improving graph-based abstract code generation will increase the productivity of users of the tools which rely on graph-based code, yet it may also cause job loss in particular fields where automation becomes predominant. It is our hope that the tools our work inspires will be designed to assist and empower human users, not replace them.

\begin{ack}

The first author conducted this research during her internship with Ramen VR. All of the authors' sincerest thanks to our dedicated and talented colleagues for their trust and support. All funding for this work is received from Ramen VR.
% Use unnumbered first level headings for the acknowledgments. All acknowledgments
% go at the end of the paper before the list of references. Moreover, you are required to declare
% funding (financial activities supporting the submitted work) and competing interests (related financial activities outside the submitted work).
% More information about this disclosure can be found at: \url{https://neurips.cc/Conferences/2025/PaperInformation/FundingDisclosure}.

% Do {\bf not} include this section in the anonymized submission, only in the final paper. You can use the \texttt{ack} environment provided in the style file to automatically hide this section in the anonymized submission.
\end{ack}

\bibliography{references}

%%%%%%%%%%%%%%%%%%%%%%%%%%%%%%%%%%%%%%%%%%%%%%%%%%%%%%%%%%%%

\appendix

\section{All ScratchTest Blocks} \label{a:all-blocks}
\begin{enumerate}
    \item WhenFlagClicked
    \item WhenKeyPressed
    \item MoveSteps
    \item TurnRight
    \item TurnLeft
    \item GoToRandom
    \item GotoXY
    \item GlideToRandom
    \item GlideToXY
    \item PointInDirection
    \item ChangeXBy
    \item SetXTo
    \item ChangeYBy
    \item SetYTo
    \item XPosition
    \item YPosition
    \item Say
    \item SayForSecs
    \item Think
    \item ThinkForSecs
    \item ChangeSizeBy
    \item SetSizeTo
    \item Wait
    \item Repeat
    \item Forever
    \item If
    \item IfElse
    \item WaitUntil
    \item RepeatUntil
    \item Stop
    \item KeyPressed
    \item MouseDown
    \item Add
    \item Subtract
    \item Multiply
    \item Divide
    \item Random
    \item GreaterThan
    \item LessThan
    \item Equals
    \item And
    \item Or
    \item Not
    \item Join
    \item LetterOf
    \item LengthOf
    \item Contains
    \item Mod
    \item Round
    \item MathFunction
    \item SetVariable
    \item ChangeVariableBy
    \item GetVariable
\end{enumerate}

\section{LLM Configurations} \label{a:llm-config}

\subsection{gpt-oss-120b}

\begin{table}[H]
  \centering
  \begin{tabular}{ll}
    \toprule
    Attribute              & Value \\
    \midrule
    Model           & openai/gpt-oss-120b \\
    Temperature     & 1 \\
    Max Completion Tokens     & 8192 \\
    Top P     & 1 \\
    Reasoning Effort     & Medium \\
    \bottomrule
  \end{tabular}
\end{table}

\subsection{qwen3-32b}

\begin{table}[H]
  \centering
  \begin{tabular}{ll}
    \toprule
    Attribute              & Value \\
    \midrule
    Model           & qwen/qwen3-32b \\
    Temperature     & 1 \\
    Max Completion Tokens     & 8192 \\
    Top P     & 1 \\
    Reasoning Effort     & Default \\
    \bottomrule
  \end{tabular}
\end{table}

\subsection{deepseek-r1-distill-llama-70b}

\begin{table}[H]
  \centering
  \begin{tabular}{ll}
    \toprule
    Attribute              & Value \\
    \midrule
    Model           & deepseek-r1-distill-llama-70b \\
    Temperature     & 1 \\
    Max Completion Tokens     & 8192 \\
    Top P     & 1 \\
    \bottomrule
  \end{tabular}
\end{table}

\subsection{llama-3.3-70b-versatile}

\begin{table}[H]
  \centering
  \begin{tabular}{ll}
    \toprule
    Attribute              & Value \\
    \midrule
    Model           & llama-3.3-70b-versatile \\
    Temperature     & 1 \\
    Max Completion Tokens     & 8192 \\
    Top P     & 1 \\
    \bottomrule
  \end{tabular}
\end{table}

\section{LLM System Prompt} \label{a:llm-system}

\#\# Goal

You are an expert software engineer extremely proficient in programming using a graph-based, visual programming language. You will be provided with the following information:

**User Query**: The overall functionality that you are required to fulfill through calling pre-declared functions.

Your task is to analyze the **User Query** and generate a connected graph that fulfills the requested functionality in **User Query**. All necessary nodes and connections MUST be included in your output. Each node can be instantiated in the graph more than once.

\#\# Node Reference (Complete Form)

Every possible node in their complete form:

\{reference\_nodes\}

\#\# Output Format

Your output must be a JSON object in the following format:
\{``nodes'': Dict[str, Node], ``edges'': List[Edge]\}

The keys of the `nodes' dictionary are NODEIDs, which are unique string identifiers such that no two nodes can have the same NODEID. A NODEID must contain only letters (a-z, A-Z) and be prefixed with ``node\_''.

The Node type is
\{``name'': NODENAME, ``value'': any|None\}
The value field of a Node type is always None except for when it is for a constant, in which case its NODENAME is ``Constant'' and its value is the value it is supposed to hold (e.g. 10, ``Hello'', True).

The Edge type is
\{``outNodeID'': str, ``outPortID'': str, ``inNodeID'': str, ``inPortID'': str\}
outNodeID is the NODEID of the node of which you are using one of the outPorts. inNodeID is the NODEID of the node of which you are using one of the inPorts. Connecting constants to a node should be done like the following example: ``moving 10 steps'' means having the edge \{``outNodeID'': IDOfAConstantNode, ``outPortID'': ``'', ``inNodeID'': ``IDOfAMoveStepsNode'', ``inPortID'': ``STEPS''\}. A ``field'' of a node can be considered an inPort for the purposes of an Edge. An outPort connects to an inPort; an outPort cannot connect to another outPort, and an inPort cannot connect to another inPort.

\section{LLM System Prompt: Alternative Output Graph Representation} \label{a:llm-system-alt}

\#\# Goal

You are an expert software engineer extremely proficient in programming using a graph-based, visual programming language. You will be provided with the following information:

**User Query**: The overall functionality that you are required to fulfill through calling pre-declared functions.

Your task is to analyze the **User Query** and generate a connected graph that fulfills the requested functionality in **User Query**. All necessary nodes and connections MUST be included in your output. Each node can be instantiated in the graph more than once.

\#\# Node Reference (Complete Form)

Every possible node in their complete form:

\{reference\_nodes\}

\#\# Output Format

Your output must be a JSON object in the following format:
\{[key: NODEID]: {``nodeName'': NODENAME, ``value'': any|None, ``edges'': List[Edge]}\}

NODEIDs are unique string identifiers such that no two nodes can have the same NODEID. A NODEID must contain only letters (a-z, A-Z) and be prefixed with ``node\_''. The value field of a Node type is always None except for when it is for a constant, in which case its NODENAME is ``Constant'' and its value is the value it is supposed to hold (e.g. 10, ``Hello'', True).

The Edge type is
\{``portID'': str, ``otherNodeID'': OTHERNODEID\}

OTHERNODEID is the NODEID this current node is connected to via the port with ID that matches the one specified in portID. An outPort connects to an inPort; an outPort cannot connect to another outPort, and an inPort cannot connect to another inPort. A Constant node has only one port, and it is an outPort: VALUE. A ``field'' of a node can be considered an inPort for the purposes of an Edge. Port connections must be defined in the ``edges'' list for both the to and from nodes.

\section{Granular Per-Prompt Results Breakdown: Reference Node Representation} \label{a:rnbreakdown}

The following results are ordered by prompt number. The prompts are:

\begin{enumerate}
    \item When the green flag is clicked, continuously move in a square pattern until the user presses the space key.
    \item When the ``r'' key is pressed, set size to a random number between 50 and 150.
    \item When the ``s'' key is pressed, say a secret password made of two random letters and three random numbers.
    \item When the green flag is clicked, simulate a loading bar by incrementally increasing a progress variable and displaying it.
    \item When the green flag is clicked, repeat until key ``q'' is pressed: move randomly.
    \item When the green flag is clicked, move forward by 50 steps and rotate right 45 degrees, repeating forever.
    \item When the ``down arrow'' key is pressed, decrease the sprite's size by 10.
    \item When the green flag is clicked, repeat until X is greater than 150: move 10 steps.
    \item When the green flag is clicked, join ``X is '' with current X position and say it.
    \item When the ``d'' key is pressed, set X to 30 and Y to 6.
    \item When the green flag is clicked, simulate a spiral by increasing steps and turning each time.
    \item When the green flag is clicked, if size is greater than 100, say ``Too big!''.
    \item When the green flag is clicked, say a sentence constructed from 3 random words.
    \item When the ``r'' key is pressed, simulate a die roll and say the result.
    \item When the green flag is clicked, count up for 5s from 0 to 100, saying each number.
    \item When the green flag is clicked, simulate a heartbeat by growing and shrinking repeatedly.
    \item When the ``left arrow'' key is pressed, change X by -20 and say ``Going left!''
    \item When the green flag is clicked, wait until the ``a'' key is pressed, then jump to (0, 0).
    \item When the green flag is clicked, alternate between saying ``Tick'' and ``Tock'' forever.
    \item When the ``x'' key is pressed, say a randomly selected word from a list.
\end{enumerate}

The results are logged using the following symbols: \textcolor{green}{$\checkmark$} means correct, \textcolor{red}{$\times$} means incorrect logic, \textcolor{red}{E} means the generation resulted in an error.

\subsection{Run 1}

\begin{table}[H]
  \centering
  \begin{tabular}{lc}
    \toprule
    Name                  & Results \\
    \midrule
    No Types              & \textcolor{red}{$\times$}\textcolor{green}{$\checkmark$}\textcolor{green}{$\checkmark$}\textcolor{red}{E}\textcolor{green}{$\checkmark$} \textcolor{green}{$\checkmark$}\textcolor{green}{$\checkmark$}\textcolor{green}{$\checkmark$}\textcolor{green}{$\checkmark$}\textcolor{green}{$\checkmark$} \textcolor{red}{$\times$}\textcolor{red}{E}\textcolor{red}{$\times$}\textcolor{green}{$\checkmark$}\textcolor{red}{$\times$} \textcolor{red}{E}\textcolor{green}{$\checkmark$}\textcolor{green}{$\checkmark$}\textcolor{green}{$\checkmark$}\textcolor{red}{E}  \\
    Extra Description     & \textcolor{green}{$\checkmark$}\textcolor{green}{$\checkmark$}\textcolor{green}{$\checkmark$}\textcolor{red}{$\times$}\textcolor{green}{$\checkmark$} \textcolor{green}{$\checkmark$}\textcolor{green}{$\checkmark$}\textcolor{green}{$\checkmark$}\textcolor{green}{$\checkmark$}\textcolor{green}{$\checkmark$} \textcolor{red}{$\times$}\textcolor{red}{E}\textcolor{green}{$\checkmark$}\textcolor{green}{$\checkmark$}\textcolor{red}{E} \textcolor{red}{E}\textcolor{green}{$\checkmark$}\textcolor{green}{$\checkmark$}\textcolor{green}{$\checkmark$}\textcolor{red}{$\times$}  \\
    \textbf{Proposed}     & \textcolor{red}{$\times$}\textcolor{green}{$\checkmark$}\textcolor{green}{$\checkmark$}\textcolor{red}{E}\textcolor{green}{$\checkmark$} \textcolor{green}{$\checkmark$}\textcolor{green}{$\checkmark$}\textcolor{green}{$\checkmark$}\textcolor{green}{$\checkmark$}\textcolor{green}{$\checkmark$} \textcolor{green}{$\checkmark$}\textcolor{red}{E}\textcolor{green}{$\checkmark$}\textcolor{green}{$\checkmark$}\textcolor{green}{$\checkmark$} \textcolor{green}{$\checkmark$}\textcolor{green}{$\checkmark$}\textcolor{green}{$\checkmark$}\textcolor{green}{$\checkmark$}\textcolor{red}{E} \\
    \bottomrule
  \end{tabular}
\end{table}

\subsection{Run 2}

\begin{table}[H]
  \centering
  \begin{tabular}{lc}
    \toprule
    Name                  & Results \\
    \midrule
    No Types              & \textcolor{red}{$\times$}\textcolor{green}{$\checkmark$}\textcolor{green}{$\checkmark$}\textcolor{green}{$\checkmark$}\textcolor{green}{$\checkmark$} \textcolor{green}{$\checkmark$}\textcolor{green}{$\checkmark$}\textcolor{green}{$\checkmark$}\textcolor{green}{$\checkmark$}\textcolor{green}{$\checkmark$} \textcolor{green}{$\checkmark$}\textcolor{red}{E}\textcolor{green}{$\checkmark$}\textcolor{red}{$\times$}\textcolor{red}{$\times$} \textcolor{red}{E}\textcolor{green}{$\checkmark$}\textcolor{green}{$\checkmark$}\textcolor{green}{$\checkmark$}\textcolor{red}{$\times$} \\
    Extra Description     & \textcolor{green}{$\checkmark$}\textcolor{green}{$\checkmark$}\textcolor{green}{$\checkmark$}\textcolor{green}{$\checkmark$}\textcolor{green}{$\checkmark$} \textcolor{green}{$\checkmark$}\textcolor{green}{$\checkmark$}\textcolor{green}{$\checkmark$}\textcolor{green}{$\checkmark$}\textcolor{green}{$\checkmark$} \textcolor{green}{$\checkmark$}\textcolor{red}{E}\textcolor{green}{$\checkmark$}\textcolor{green}{$\checkmark$}\textcolor{green}{$\checkmark$} \textcolor{red}{E}\textcolor{green}{$\checkmark$}\textcolor{green}{$\checkmark$}\textcolor{red}{$\times$}\textcolor{red}{E} \\
    \textbf{Proposed}     & \textcolor{green}{$\checkmark$}\textcolor{green}{$\checkmark$}\textcolor{green}{$\checkmark$}\textcolor{red}{$\times$}\textcolor{green}{$\checkmark$} \textcolor{green}{$\checkmark$}\textcolor{red}{$\times$}\textcolor{green}{$\checkmark$}\textcolor{green}{$\checkmark$}\textcolor{green}{$\checkmark$} \textcolor{green}{$\checkmark$}\textcolor{red}{E}\textcolor{green}{$\checkmark$}\textcolor{red}{$\times$}\textcolor{red}{$\times$} \textcolor{red}{E}\textcolor{green}{$\checkmark$}\textcolor{green}{$\checkmark$}\textcolor{green}{$\checkmark$}\textcolor{green}{$\checkmark$} \\
    \bottomrule
  \end{tabular}
\end{table}

\subsection{Run 3}

\begin{table}[H]
  \centering
  \begin{tabular}{lc}
    \toprule
    Name                  & Results \\
    \midrule
    No Types              & \textcolor{red}{E}\textcolor{green}{$\checkmark$}\textcolor{red}{$\times$}\textcolor{green}{$\checkmark$}\textcolor{green}{$\checkmark$} \textcolor{green}{$\checkmark$}\textcolor{green}{$\checkmark$}\textcolor{green}{$\checkmark$}\textcolor{green}{$\checkmark$}\textcolor{green}{$\checkmark$} \textcolor{red}{$\times$}\textcolor{red}{E}\textcolor{red}{E}\textcolor{green}{$\checkmark$}\textcolor{red}{$\times$} \textcolor{red}{E}\textcolor{green}{$\checkmark$}\textcolor{green}{$\checkmark$}\textcolor{green}{$\checkmark$}\textcolor{red}{E} \\
    Extra Description     & \textcolor{green}{$\checkmark$}\textcolor{green}{$\checkmark$}\textcolor{red}{$\times$}\textcolor{green}{$\checkmark$}\textcolor{green}{$\checkmark$} \textcolor{green}{$\checkmark$}\textcolor{green}{$\checkmark$}\textcolor{green}{$\checkmark$}\textcolor{green}{$\checkmark$}\textcolor{green}{$\checkmark$} \textcolor{red}{$\times$}\textcolor{red}{E}\textcolor{red}{E}\textcolor{green}{$\checkmark$}\textcolor{red}{E} \textcolor{green}{$\checkmark$}\textcolor{green}{$\checkmark$}\textcolor{green}{$\checkmark$}\textcolor{green}{$\checkmark$}\textcolor{red}{E} \\
    \textbf{Proposed}     & \textcolor{red}{E}\textcolor{green}{$\checkmark$}\textcolor{green}{$\checkmark$}\textcolor{red}{$\times$}\textcolor{green}{$\checkmark$} \textcolor{green}{$\checkmark$}\textcolor{green}{$\checkmark$}\textcolor{green}{$\checkmark$}\textcolor{green}{$\checkmark$}\textcolor{green}{$\checkmark$} \textcolor{green}{$\checkmark$}\textcolor{red}{E}\textcolor{green}{$\checkmark$}\textcolor{green}{$\checkmark$}\textcolor{red}{$\times$} \textcolor{green}{$\checkmark$}\textcolor{green}{$\checkmark$}\textcolor{green}{$\checkmark$}\textcolor{green}{$\checkmark$}\textcolor{red}{E} \\
    \bottomrule
  \end{tabular}
\end{table}

\subsection{Run 4}

\begin{table}[H]
  \centering
  \begin{tabular}{lc}
    \toprule
    Name                  & Results \\
    \midrule
    No Types              & \textcolor{green}{$\checkmark$}\textcolor{green}{$\checkmark$}\textcolor{green}{$\checkmark$}\textcolor{red}{$\times$}\textcolor{green}{$\checkmark$} \textcolor{green}{$\checkmark$}\textcolor{red}{E}\textcolor{green}{$\checkmark$}\textcolor{green}{$\checkmark$}\textcolor{green}{$\checkmark$} \textcolor{green}{$\checkmark$}\textcolor{red}{E}\textcolor{green}{$\checkmark$}\textcolor{red}{$\times$}\textcolor{green}{$\checkmark$} \textcolor{red}{E}\textcolor{green}{$\checkmark$}\textcolor{green}{$\checkmark$}\textcolor{green}{$\checkmark$}\textcolor{green}{$\checkmark$} \\
    Extra Description     & \textcolor{green}{$\checkmark$}\textcolor{green}{$\checkmark$}\textcolor{red}{E}\textcolor{green}{$\checkmark$}\textcolor{green}{$\checkmark$} \textcolor{green}{$\checkmark$}\textcolor{green}{$\checkmark$}\textcolor{green}{$\checkmark$}\textcolor{green}{$\checkmark$}\textcolor{green}{$\checkmark$} \textcolor{green}{$\checkmark$}\textcolor{red}{E}\textcolor{green}{$\checkmark$}\textcolor{green}{$\checkmark$}\textcolor{green}{$\checkmark$} \textcolor{red}{E}\textcolor{green}{$\checkmark$}\textcolor{green}{$\checkmark$}\textcolor{green}{$\checkmark$}\textcolor{green}{$\checkmark$} \\
    \textbf{Proposed}     & \textcolor{green}{$\checkmark$}\textcolor{green}{$\checkmark$}\textcolor{green}{$\checkmark$}\textcolor{red}{$\times$}\textcolor{green}{$\checkmark$} \textcolor{green}{$\checkmark$}\textcolor{red}{E}\textcolor{green}{$\checkmark$}\textcolor{green}{$\checkmark$}\textcolor{green}{$\checkmark$} \textcolor{green}{$\checkmark$}\textcolor{green}{$\checkmark$}\textcolor{green}{$\checkmark$}\textcolor{green}{$\checkmark$}\textcolor{red}{$\times$} \textcolor{green}{$\checkmark$}\textcolor{green}{$\checkmark$}\textcolor{green}{$\checkmark$}\textcolor{green}{$\checkmark$}\textcolor{red}{E} \\
    \bottomrule
  \end{tabular}
\end{table}

\subsection{Run 5}

\begin{table}[H]
  \centering
  \begin{tabular}{lc}
    \toprule
    Name                  & Results \\
    \midrule
    No Types              & \textcolor{red}{E}\textcolor{green}{$\checkmark$}\textcolor{green}{$\checkmark$}\textcolor{red}{$\times$}\textcolor{green}{$\checkmark$} \textcolor{green}{$\checkmark$}\textcolor{green}{$\checkmark$}\textcolor{green}{$\checkmark$}\textcolor{green}{$\checkmark$}\textcolor{green}{$\checkmark$} \textcolor{red}{$\times$}\textcolor{red}{E}\textcolor{red}{$\times$}\textcolor{green}{$\checkmark$}\textcolor{red}{$\times$} \textcolor{red}{E}\textcolor{green}{$\checkmark$}\textcolor{green}{$\checkmark$}\textcolor{green}{$\checkmark$}\textcolor{red}{E} \\
    Extra Description     & \textcolor{green}{$\checkmark$}\textcolor{green}{$\checkmark$}\textcolor{green}{$\checkmark$}\textcolor{red}{$\times$}\textcolor{green}{$\checkmark$} \textcolor{green}{$\checkmark$}\textcolor{green}{$\checkmark$}\textcolor{green}{$\checkmark$}\textcolor{green}{$\checkmark$}\textcolor{green}{$\checkmark$} \textcolor{red}{$\times$}\textcolor{red}{E}\textcolor{red}{E}\textcolor{green}{$\checkmark$}\textcolor{green}{$\checkmark$} \textcolor{red}{E}\textcolor{green}{$\checkmark$}\textcolor{green}{$\checkmark$}\textcolor{red}{$\times$}\textcolor{red}{E} \\
    \textbf{Proposed}     & \textcolor{green}{$\checkmark$}\textcolor{green}{$\checkmark$}\textcolor{green}{$\checkmark$}\textcolor{red}{$\times$}\textcolor{green}{$\checkmark$} \textcolor{green}{$\checkmark$}\textcolor{red}{E}\textcolor{green}{$\checkmark$}\textcolor{green}{$\checkmark$}\textcolor{green}{$\checkmark$} \textcolor{red}{$\times$}\textcolor{red}{E}\textcolor{green}{$\checkmark$}\textcolor{green}{$\checkmark$}\textcolor{red}{$\times$} \textcolor{red}{E}\textcolor{green}{$\checkmark$}\textcolor{green}{$\checkmark$}\textcolor{green}{$\checkmark$}\textcolor{green}{$\checkmark$} \\
    \bottomrule
  \end{tabular}
\end{table}

\section{Granular Per-Prompt Results Breakdown: Output Graph Representation} \label{a:ogbreakdown}

The following results are ordered by prompt number. The prompts are the same as in Appendix \ref{a:rnbreakdown}.

The results are logged using the following symbols: \textcolor{green}{$\checkmark$} means correct, \textcolor{red}{$\times$} means incorrect logic, \textcolor{red}{E} means the generation resulted in an error.

\subsection{Run 1}

\begin{table}[H]
  \centering
  \begin{tabular}{lc}
    \toprule
    Name                  & Results \\
    \midrule
    Alternative     & \textcolor{red}{E}\textcolor{green}{$\checkmark$}\textcolor{red}{$\times$}\textcolor{red}{E}\textcolor{green}{$\checkmark$} \textcolor{green}{$\checkmark$}\textcolor{green}{$\checkmark$}\textcolor{red}{E}\textcolor{green}{$\checkmark$}\textcolor{red}{$\times$} \textcolor{red}{$\times$}\textcolor{red}{E}\textcolor{green}{$\checkmark$}\textcolor{red}{E}\textcolor{green}{$\checkmark$} \textcolor{red}{E}\textcolor{green}{$\checkmark$}\textcolor{green}{$\checkmark$}\textcolor{green}{$\checkmark$}\textcolor{red}{E}  \\
    \textbf{Proposed}     & \textcolor{red}{$\times$}\textcolor{green}{$\checkmark$}\textcolor{green}{$\checkmark$}\textcolor{red}{E}\textcolor{green}{$\checkmark$} \textcolor{green}{$\checkmark$}\textcolor{green}{$\checkmark$}\textcolor{green}{$\checkmark$}\textcolor{green}{$\checkmark$}\textcolor{green}{$\checkmark$} \textcolor{green}{$\checkmark$}\textcolor{red}{E}\textcolor{green}{$\checkmark$}\textcolor{green}{$\checkmark$}\textcolor{green}{$\checkmark$} \textcolor{green}{$\checkmark$}\textcolor{green}{$\checkmark$}\textcolor{green}{$\checkmark$}\textcolor{green}{$\checkmark$}\textcolor{red}{E} \\
    \bottomrule
  \end{tabular}
\end{table}

\subsection{Run 2}

\begin{table}[H]
  \centering
  \begin{tabular}{lc}
    \toprule
    Name                  & Results \\
    \midrule
    Alternative             & \textcolor{red}{E}\textcolor{green}{$\checkmark$}\textcolor{green}{$\checkmark$}\textcolor{red}{E}\textcolor{red}{E} \textcolor{red}{E}\textcolor{green}{$\checkmark$}\textcolor{red}{E}\textcolor{green}{$\checkmark$}\textcolor{green}{$\checkmark$} \textcolor{red}{E}\textcolor{red}{E}\textcolor{green}{$\checkmark$}\textcolor{green}{$\checkmark$}\textcolor{red}{E} \textcolor{green}{$\checkmark$}\textcolor{green}{$\checkmark$}\textcolor{red}{$\times$}\textcolor{red}{E}\textcolor{red}{E} \\
    \textbf{Proposed}     & \textcolor{green}{$\checkmark$}\textcolor{green}{$\checkmark$}\textcolor{green}{$\checkmark$}\textcolor{red}{$\times$}\textcolor{green}{$\checkmark$} \textcolor{green}{$\checkmark$}\textcolor{red}{$\times$}\textcolor{green}{$\checkmark$}\textcolor{green}{$\checkmark$}\textcolor{green}{$\checkmark$} \textcolor{green}{$\checkmark$}\textcolor{red}{E}\textcolor{green}{$\checkmark$}\textcolor{red}{$\times$}\textcolor{red}{$\times$} \textcolor{red}{E}\textcolor{green}{$\checkmark$}\textcolor{green}{$\checkmark$}\textcolor{green}{$\checkmark$}\textcolor{green}{$\checkmark$} \\
    \bottomrule
  \end{tabular}
\end{table}

\subsection{Run 3}

\begin{table}[H]
  \centering
  \begin{tabular}{lc}
    \toprule
    Name                  & Results \\
    \midrule
    Alternative              & \textcolor{red}{E}\textcolor{green}{$\checkmark$}\textcolor{red}{E}\textcolor{red}{E}\textcolor{green}{$\checkmark$} \textcolor{green}{$\checkmark$}\textcolor{green}{$\checkmark$}\textcolor{red}{E}\textcolor{green}{$\checkmark$}\textcolor{green}{$\checkmark$} \textcolor{red}{E}\textcolor{red}{E}\textcolor{green}{$\checkmark$}\textcolor{green}{$\checkmark$}\textcolor{red}{$\times$} \textcolor{red}{E}\textcolor{green}{$\checkmark$}\textcolor{red}{$\times$}\textcolor{red}{E}\textcolor{green}{$\checkmark$} \\
    \textbf{Proposed}     & \textcolor{red}{E}\textcolor{green}{$\checkmark$}\textcolor{green}{$\checkmark$}\textcolor{red}{$\times$}\textcolor{green}{$\checkmark$} \textcolor{green}{$\checkmark$}\textcolor{green}{$\checkmark$}\textcolor{green}{$\checkmark$}\textcolor{green}{$\checkmark$}\textcolor{green}{$\checkmark$} \textcolor{green}{$\checkmark$}\textcolor{red}{E}\textcolor{green}{$\checkmark$}\textcolor{green}{$\checkmark$}\textcolor{red}{$\times$} \textcolor{green}{$\checkmark$}\textcolor{green}{$\checkmark$}\textcolor{green}{$\checkmark$}\textcolor{green}{$\checkmark$}\textcolor{red}{E} \\
    \bottomrule
  \end{tabular}
\end{table}

\subsection{Run 4}

\begin{table}[H]
  \centering
  \begin{tabular}{lc}
    \toprule
    Name                  & Results \\
    \midrule
    Alternative              & \textcolor{red}{E}\textcolor{green}{$\checkmark$}\textcolor{green}{$\checkmark$}\textcolor{red}{E}\textcolor{red}{E} \textcolor{green}{$\checkmark$}\textcolor{green}{$\checkmark$}\textcolor{green}{$\checkmark$}\textcolor{green}{$\checkmark$}\textcolor{green}{$\checkmark$} \textcolor{red}{E}\textcolor{red}{E}\textcolor{green}{$\checkmark$}\textcolor{green}{$\checkmark$}\textcolor{red}{E} \textcolor{red}{E}\textcolor{green}{$\checkmark$}\textcolor{green}{$\checkmark$}\textcolor{red}{E}\textcolor{red}{E} \\
    \textbf{Proposed}     & \textcolor{green}{$\checkmark$}\textcolor{green}{$\checkmark$}\textcolor{green}{$\checkmark$}\textcolor{red}{$\times$}\textcolor{green}{$\checkmark$} \textcolor{green}{$\checkmark$}\textcolor{red}{E}\textcolor{green}{$\checkmark$}\textcolor{green}{$\checkmark$}\textcolor{green}{$\checkmark$} \textcolor{green}{$\checkmark$}\textcolor{green}{$\checkmark$}\textcolor{green}{$\checkmark$}\textcolor{green}{$\checkmark$}\textcolor{red}{$\times$} \textcolor{green}{$\checkmark$}\textcolor{green}{$\checkmark$}\textcolor{green}{$\checkmark$}\textcolor{green}{$\checkmark$}\textcolor{red}{E} \\
    \bottomrule
  \end{tabular}
\end{table}

\subsection{Run 5}

\begin{table}[H]
  \centering
  \begin{tabular}{lc}
    \toprule
    Name                  & Results \\
    \midrule
    Alternative           & \textcolor{red}{E}\textcolor{green}{$\checkmark$}\textcolor{green}{$\checkmark$}\textcolor{red}{E}\textcolor{red}{E} \textcolor{red}{E}\textcolor{green}{$\checkmark$}\textcolor{green}{$\checkmark$}\textcolor{green}{$\checkmark$}\textcolor{green}{$\checkmark$} \textcolor{red}{$\times$}\textcolor{red}{E}\textcolor{green}{$\checkmark$}\textcolor{red}{$\times$}\textcolor{red}{$\times$} \textcolor{red}{E}\textcolor{green}{$\checkmark$}\textcolor{green}{$\checkmark$}\textcolor{red}{E}\textcolor{red}{E} \\
    \textbf{Proposed}     & \textcolor{green}{$\checkmark$}\textcolor{green}{$\checkmark$}\textcolor{green}{$\checkmark$}\textcolor{red}{$\times$}\textcolor{green}{$\checkmark$} \textcolor{green}{$\checkmark$}\textcolor{red}{E}\textcolor{green}{$\checkmark$}\textcolor{green}{$\checkmark$}\textcolor{green}{$\checkmark$} \textcolor{red}{$\times$}\textcolor{red}{E}\textcolor{green}{$\checkmark$}\textcolor{green}{$\checkmark$}\textcolor{red}{$\times$} \textcolor{red}{E}\textcolor{green}{$\checkmark$}\textcolor{green}{$\checkmark$}\textcolor{green}{$\checkmark$}\textcolor{green}{$\checkmark$} \\
    \bottomrule
  \end{tabular}
\end{table}

\section{Evaluation for Other LLMs} \label{a:other-llms}

The following results are ordered by prompt number. The prompts are the same as in Appendix \ref{a:rnbreakdown}.

The results are logged using the following symbols: \textcolor{green}{$\checkmark$} means correct, \textcolor{red}{$\times$} means incorrect logic, \textcolor{red}{E} means the generation resulted in an error.

\subsection{qwen3-32b}

\begin{table}[H]
  \caption{ScratchTest results for reference node representation ablations.}
  \centering
  \begin{tabular}{lc}
    \toprule
    Name                  & Results \\
    \midrule
    No Types              & \textcolor{red}{E}\textcolor{red}{$\times$}\textcolor{red}{E}\textcolor{red}{E}\textcolor{red}{E} \textcolor{red}{E}\textcolor{red}{$\times$}\textcolor{green}{$\checkmark$}\textcolor{green}{$\checkmark$}\textcolor{red}{$\times$} \textcolor{red}{E}\textcolor{red}{E}\textcolor{red}{$\times$}\textcolor{red}{$\times$}\textcolor{red}{E} \textcolor{red}{E}\textcolor{red}{$\times$}\textcolor{green}{$\checkmark$}\textcolor{red}{E}\textcolor{red}{$\times$}  \\

    Extra Description     & \textcolor{red}{E}\textcolor{red}{E}\textcolor{red}{$\times$}\textcolor{red}{E}\textcolor{green}{$\checkmark$} \textcolor{red}{E}\textcolor{red}{$\times$}\textcolor{red}{E}\textcolor{green}{$\checkmark$}\textcolor{red}{$\times$} \textcolor{red}{$\times$}\textcolor{red}{E}\textcolor{red}{E}\textcolor{red}{$\times$}\textcolor{red}{E} \textcolor{red}{$\times$}\textcolor{red}{E}\textcolor{red}{E}\textcolor{red}{E}\textcolor{red}{$\times$}  \\

    \textbf{Proposed}     & \textcolor{red}{E}\textcolor{red}{$\times$}\textcolor{green}{$\checkmark$}\textcolor{red}{E}\textcolor{red}{$\times$} \textcolor{red}{$\times$}\textcolor{red}{$\times$}\textcolor{green}{$\checkmark$}\textcolor{green}{$\checkmark$}\textcolor{red}{$\times$} \textcolor{red}{E}\textcolor{red}{E}\textcolor{green}{$\checkmark$}\textcolor{red}{$\times$}\textcolor{red}{E} \textcolor{red}{E}\textcolor{red}{$\times$}\textcolor{green}{$\checkmark$}\textcolor{red}{$\times$}\textcolor{red}{E} \\
    \bottomrule
  \end{tabular}
\end{table}

\begin{table}[H]
  \caption{ScratchTest results for output graph representation ablations.}
  \centering
  \begin{tabular}{lc}
    \toprule
    Name                  & Results \\
    \midrule
    Alternative           & \textcolor{red}{E}\textcolor{red}{E}\textcolor{red}{E}\textcolor{red}{E}\textcolor{red}{E} \textcolor{red}{E}\textcolor{red}{E}\textcolor{red}{E}\textcolor{red}{E}\textcolor{red}{$\times$} \textcolor{red}{E}\textcolor{red}{E}\textcolor{red}{E}\textcolor{red}{E}\textcolor{red}{E} \textcolor{red}{E}\textcolor{red}{E}\textcolor{red}{E}\textcolor{red}{E}\textcolor{red}{E} \\
    \textbf{Proposed}     & \textcolor{red}{E}\textcolor{red}{$\times$}\textcolor{green}{$\checkmark$}\textcolor{red}{E}\textcolor{red}{$\times$} \textcolor{red}{$\times$}\textcolor{red}{$\times$}\textcolor{green}{$\checkmark$}\textcolor{green}{$\checkmark$}\textcolor{red}{$\times$} \textcolor{red}{E}\textcolor{red}{E}\textcolor{green}{$\checkmark$}\textcolor{red}{$\times$}\textcolor{red}{E} \textcolor{red}{E}\textcolor{red}{$\times$}\textcolor{green}{$\checkmark$}\textcolor{red}{$\times$}\textcolor{red}{E} \\
    \bottomrule
  \end{tabular}
\end{table}

\subsection{deepseek-r1-distill-llama-70b}

\begin{table}[H]
  \caption{ScratchTest results for reference node representation ablations.}
  \centering
  \begin{tabular}{lc}
    \toprule
    Name                  & Results \\
    \midrule
    No Types              & \textcolor{red}{E}\textcolor{red}{$\times$}\textcolor{red}{E}\textcolor{red}{E}\textcolor{red}{$\times$} \textcolor{red}{$\times$}\textcolor{red}{$\times$}\textcolor{red}{E}\textcolor{green}{$\checkmark$}\textcolor{green}{$\checkmark$} \textcolor{red}{E}\textcolor{red}{E}\textcolor{green}{$\checkmark$}\textcolor{red}{E}\textcolor{red}{E} \textcolor{red}{E}\textcolor{red}{$\times$}\textcolor{green}{$\checkmark$}\textcolor{red}{E}\textcolor{red}{E}  \\

    Extra Description     & \textcolor{red}{E}\textcolor{red}{E}\textcolor{red}{$\times$}\textcolor{red}{E}\textcolor{green}{$\checkmark$} \textcolor{red}{E}\textcolor{red}{E}\textcolor{green}{$\checkmark$}\textcolor{green}{$\checkmark$}\textcolor{red}{$\times$} \textcolor{red}{E}\textcolor{red}{E}\textcolor{red}{E}\textcolor{red}{$\times$}\textcolor{red}{E} \textcolor{red}{$\times$}\textcolor{red}{$\times$}\textcolor{green}{$\checkmark$}\textcolor{red}{E}\textcolor{red}{E}  \\

    \textbf{Proposed}     & \textcolor{red}{E}\textcolor{red}{$\times$}\textcolor{red}{E}\textcolor{red}{E}\textcolor{green}{$\checkmark$} \textcolor{green}{$\checkmark$}\textcolor{red}{$\times$}\textcolor{green}{$\checkmark$}\textcolor{green}{$\checkmark$}\textcolor{red}{E} \textcolor{red}{E}\textcolor{red}{E}\textcolor{red}{E}\textcolor{red}{E}\textcolor{red}{E} \textcolor{green}{$\checkmark$}\textcolor{red}{$\times$}\textcolor{green}{$\checkmark$}\textcolor{red}{E}\textcolor{red}{E} \\
    \bottomrule
  \end{tabular}
\end{table}

\begin{table}[H]
  \caption{ScratchTest results for output graph representation ablations.}
  \centering
  \begin{tabular}{lc}
    \toprule
    Name                  & Results \\
    \midrule
    Alternative           & \textcolor{red}{E}\textcolor{red}{E}\textcolor{red}{E}\textcolor{red}{E}\textcolor{red}{E} \textcolor{red}{E}\textcolor{red}{E}\textcolor{red}{E}\textcolor{green}{$\checkmark$}\textcolor{red}{E} \textcolor{red}{E}\textcolor{red}{E}\textcolor{red}{E}\textcolor{red}{E}\textcolor{red}{E} \textcolor{red}{E}\textcolor{red}{E}\textcolor{red}{E}\textcolor{red}{E}\textcolor{red}{E} \\
    \textbf{Proposed}     & \textcolor{red}{E}\textcolor{red}{$\times$}\textcolor{red}{E}\textcolor{red}{E}\textcolor{green}{$\checkmark$} \textcolor{green}{$\checkmark$}\textcolor{red}{$\times$}\textcolor{green}{$\checkmark$}\textcolor{green}{$\checkmark$}\textcolor{red}{E} \textcolor{red}{E}\textcolor{red}{E}\textcolor{red}{E}\textcolor{red}{E}\textcolor{red}{E} \textcolor{green}{$\checkmark$}\textcolor{red}{$\times$}\textcolor{green}{$\checkmark$}\textcolor{red}{E}\textcolor{red}{E} \\
    \bottomrule
  \end{tabular}
\end{table}

\subsection{llama-3.3-70b-versatile}

\begin{table}[H]
  \caption{ScratchTest results for reference node representation ablations.}
  \centering
  \begin{tabular}{lc}
    \toprule
    Name                  & Results \\
    \midrule
    No Types              & \textcolor{red}{E}\textcolor{green}{$\checkmark$}\textcolor{red}{$\times$}\textcolor{red}{E}\textcolor{red}{E} \textcolor{green}{$\checkmark$}\textcolor{red}{E}\textcolor{red}{E}\textcolor{green}{$\checkmark$}\textcolor{green}{$\checkmark$} \textcolor{red}{E}\textcolor{red}{E}\textcolor{red}{E}\textcolor{green}{$\checkmark$}\textcolor{red}{E} \textcolor{red}{E}\textcolor{red}{E}\textcolor{green}{$\checkmark$}\textcolor{red}{E}\textcolor{red}{$\times$}  \\

    Extra Description     & \textcolor{red}{E}\textcolor{green}{$\checkmark$}\textcolor{red}{E}\textcolor{red}{E}\textcolor{red}{E} \textcolor{red}{E}\textcolor{red}{E}\textcolor{green}{$\checkmark$}\textcolor{red}{E}\textcolor{red}{E} \textcolor{red}{E}\textcolor{red}{E}\textcolor{red}{$\times$}\textcolor{red}{E}\textcolor{red}{E} \textcolor{red}{E}\textcolor{green}{$\checkmark$}\textcolor{red}{E}\textcolor{red}{E}\textcolor{red}{$\times$}  \\

    \textbf{Proposed}     & \textcolor{red}{E}\textcolor{green}{$\checkmark$}\textcolor{red}{E}\textcolor{red}{E}\textcolor{red}{$\times$} \textcolor{red}{E}\textcolor{green}{$\checkmark$}\textcolor{red}{E}\textcolor{green}{$\checkmark$}\textcolor{red}{$\times$} \textcolor{red}{E}\textcolor{green}{$\checkmark$}\textcolor{green}{$\checkmark$}\textcolor{red}{E}\textcolor{green}{$\checkmark$} \textcolor{red}{E}\textcolor{green}{$\checkmark$}\textcolor{green}{$\checkmark$}\textcolor{red}{E}\textcolor{red}{E} \\
    \bottomrule
  \end{tabular}
\end{table}

\begin{table}[H]
  \caption{ScratchTest results for output graph representation ablations.}
  \centering
  \begin{tabular}{lc}
    \toprule
    Name                  & Results \\
    \midrule
    Alternative           & \textcolor{red}{E}\textcolor{red}{E}\textcolor{red}{E}\textcolor{red}{E}\textcolor{red}{E} \textcolor{red}{E}\textcolor{red}{E}\textcolor{red}{E}\textcolor{red}{E}\textcolor{red}{$\times$} \textcolor{green}{$\checkmark$}\textcolor{red}{E}\textcolor{red}{E}\textcolor{red}{E}\textcolor{red}{E} \textcolor{red}{E}\textcolor{red}{E}\textcolor{red}{E}\textcolor{red}{E}\textcolor{red}{E} \\
    \textbf{Proposed}     & \textcolor{red}{E}\textcolor{green}{$\checkmark$}\textcolor{red}{E}\textcolor{red}{E}\textcolor{red}{$\times$} \textcolor{red}{E}\textcolor{green}{$\checkmark$}\textcolor{red}{E}\textcolor{green}{$\checkmark$}\textcolor{red}{$\times$} \textcolor{red}{E}\textcolor{green}{$\checkmark$}\textcolor{green}{$\checkmark$}\textcolor{red}{E}\textcolor{green}{$\checkmark$} \textcolor{red}{E}\textcolor{green}{$\checkmark$}\textcolor{green}{$\checkmark$}\textcolor{red}{E}\textcolor{red}{E} \\
    \bottomrule
  \end{tabular}
\end{table}

\end{document}